\RequirePackage{silence}
\PassOptionsToPackage{hypertexnames=false}{hyperref}

\documentclass[a4paper,fleqn]{cas-dc}
\usepackage{soul}
\usepackage[numbers]{natbib}
\usepackage{algorithm}
\usepackage{algpseudocode}
\usepackage{pifont}

\usepackage{bm}
\usepackage{tabularx}

\definecolor{brightgreen}{rgb}{0.0, 0.5, 0.0}
\usepackage{array}

\def\tsc#1{\csdef{#1}{\textsc{\lowercase{#1}}\xspace}}
\tsc{WGM}
\tsc{QE}
\tsc{EP}
\tsc{PMS}
\tsc{BEC}
\tsc{DE}

\begin{document}

\def\floatpagepagefraction{1}
\def\textpagefraction{.001}

\shorttitle{SwinAD: Multi-stage feature reconstruction for unsupervised industrial anomaly detection}
\shortauthors{H. Ninh, C. Thai, M.X. Trang, V. Le, T.H. Le, L. Tran}

\author[1]{Huong Ninh}
\ead{25028005@vnu.edu.vn} \fnmark[1]
\author[3]{Chien Thai}[type=editor, auid=000, bioid=1, orcid=0000-0002-5098-6862]
\ead{chien.thaivan@phenikaa-uni.edu.vn} \fnmark[1] 
\author[3]{Mai Xuan Trang}[orcid=0000-0002-9666-0198]
\cormark[1]
\ead{trang.maixuan@phenikaa-uni.edu.vn}
\author[1]{Vu-Minh Le}
\ead{21020649@vnu.edu.vn}
\author[2]{Thanh Ha Le}
\ead{ltha@vnu.edu.vn}
\author[1]{Long Tran}
\ead{tqlong@vnu.edu.vn} 
\cormark[1]

\affiliation[1]{organization={Institute of Artificial Intelligence, University of Engineering and Technology, Vietnam National University}}
\affiliation[2]{organization={Human-Machine Interaction Laboratory, University of Engineering and Technology, Vietnam National University}}
\affiliation[3]{organization={Applied AI Lab, Phenikaa School of Computing, Phenikaa University}}

\fntext[1]{These authors contributed equally to this work as co first authors.}
\cortext[1]{Corresponding author.}
\title [mode = title]{SwinAD: Multi-stage feature reconstruction for unsupervised industrial anomaly detection}           

\begin{abstract}
Industrial anomaly detection aims to identify and localize defective regions without relying on exhaustive annotations of all possible defect types. Although recent unsupervised methods have achieved strong performance, most are primarily designed for single-class settings and often struggle in multi-class scenarios, where diverse normal patterns may lead to over-generalization and reduce the discriminative capability between normal and anomalous regions. In this paper, we propose \textbf{SwinAD}, a reconstruction-based framework for multi-class unsupervised anomaly detection that leverages a frozen pretrained Swin Transformer V2 encoder and a feature diversity-preserving reconstruction decoder. The hierarchical encoder provides semantically rich multi-scale features, while stage-wise bottleneck modules  with dropout prevent trivial identity mapping and encourage robust reconstruction of normal patterns. To further improve localization, we introduce a feature diversity-preserving reconstruction framework that maintains complementary reconstruction hypotheses instead of relying on a single decoding branch. The discrepancies between encoder features and the two reconstructed features are then aggregated across multiple scales to produce the final anomaly map. Experiments conducted on three industrial anomaly detection benchmarks, including MVTec AD, VisA, and Real-IAD, demonstrate that SwinAD achieves competitive image-level performance and strong pixel-level localization accuracy, with particularly notable improvements in pixel-level AP and $F_1$ on MVTec AD. These results indicate that combining hierarchical Swin features with diverse multi-scale reconstruction provides an effective and efficient solution for multi-class unsupervised anomaly detection.

\end{abstract}

\maketitle
\hfuzz=25pt
\section{Introduction}
\label{sec:intro}

Visual anomaly detection (VAD) plays an important role in intelligent inspection systems \cite{zhang2025diverse, zhang2026adaptive, li2026ipg, carrera2015detecting}, where automated defect detection must operate reliably across diverse industrial environments. In practical deployments, anomaly detection models are expected not only to identify defective products, but also to localize anomalous regions accurately while maintaining computational efficiency and scalability across multiple product categories. These requirements are particularly critical for modern industrial AI systems, where inspection pipelines must support large-scale production with limited computational resources and low inference latency. Compared with conventional supervised learning problems, anomaly detection is inherently challenging because defective samples are typically rare, highly diverse, and difficult to comprehensively annotate. In many practical scenarios, collecting sufficient defect samples for all possible abnormal conditions is either prohibitively expensive or even infeasible. Hence, recent research has increasingly focused on unsupervised learning paradigms, where models are trained exclusively on normal samples, and anomalies are identified as deviations from learned normal patterns.

\begin{figure}[pos=t]
    \centering
    \includegraphics[width=1.0\linewidth]{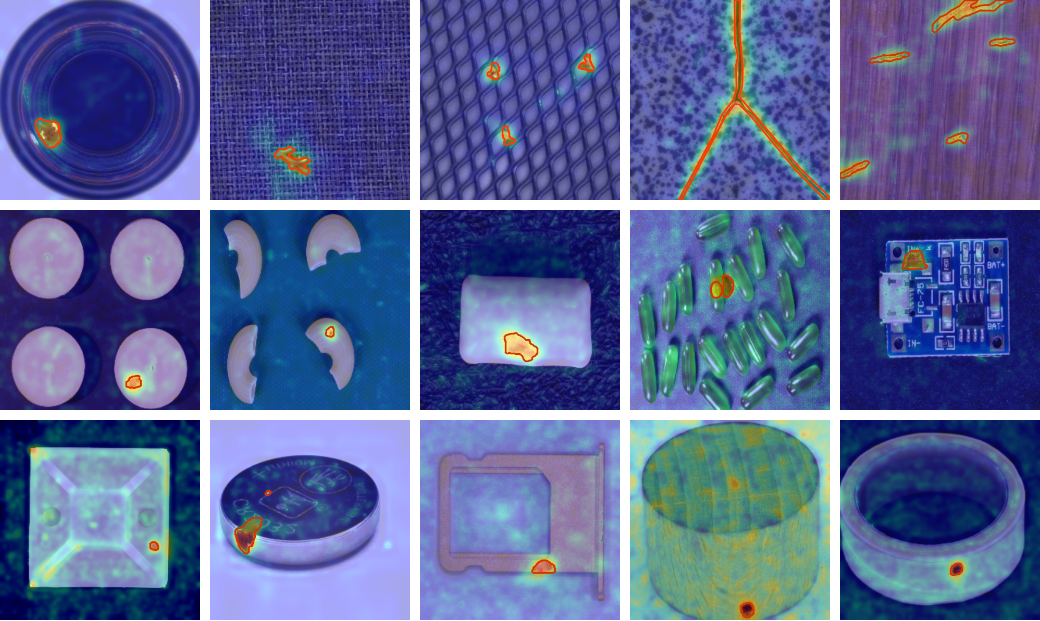}
    \caption{Examples from the MVTec AD (top), VisA (middle), and Real-IAD (bottom) datasets with SwinAD segmentation results overlaid. Orange contours indicate anomaly boundaries, while the blue-to-orange heatmaps visualize anomaly intensity for defects such as broken glass, scratches, and structural changes.}
    \label{fig:intro_samples}
\end{figure}

Existing unsupervised anomaly detection approaches are primarily developed for the Single-class Unsupervised Anomaly Detection (SUAD) setting \cite{deng2022anomaly, simplenet, zhang2023destseg}, where an independent model is trained for each object category. Although effective, this design requires high computational and memory costs when scaling to large industrial systems with numerous product types. Maintaining separate models for each category significantly increases training costs, memory consumption, deployment complexity, and system maintenance overhead. In contrast, Multi-class Unsupervised Anomaly Detection (MUAD) methods employ a single unified model across multiple categories \cite{uniad, vitad, dinomaly}. Despite its scalability advantages, MUAD is more challenging due to the diversity of normal patterns; thus, improving both detection performance and training efficiency in MUAD is a non-trivial task. Furthermore, precise localization is often more important than image-level classification alone. In practical scenarios, inaccurate anomaly boundaries may lead to incorrect defect analysis, unnecessary product rejection, or failure to identify small structural defects. Therefore, designing MUAD frameworks that simultaneously preserve localization accuracy, semantic robustness, and computational efficiency remains a critical challenge. Motivated by these challenges, our work focuses on the more practical and scalable MUAD setting.

Current unsupervised anomaly detection approaches can be divided into three main categories: Augmentation-based methods \cite{cutpaste, nsa}, Embedding-based methods \cite{padim, patchcore, wang2021student}, and Reconstruction-based methods \cite{vitad, dinomaly, he2024mambaad}. Augmentation-based approaches improve anomaly discrimination by generating synthetic defects during training. Embedding-based methods learn latent feature representations of normal samples, where anomalies can be identified through feature-space differences. Among these approaches, reconstruction-based methods have attracted significant attention due to their intuitive learning objective and strong localization capability. By measuring the difference between the input and reconstructed representations, these methods can generate effective pixel-level anomaly maps that accurately highlight defective regions. This property makes reconstruction-based approaches particularly suitable for industrial inspection tasks that require precise anomaly localization. However, existing reconstruction-based methods primarily emphasize high-level semantic representations for global anomaly discrimination, which may suppress fine-grained spatial details required for precise localization. Consequently, achieving robust localization performance while maintaining efficient inference remains an open challenge for scalable industrial anomaly detection systems.

To address the challenges of MUAD while leveraging the strengths of reconstruction-based anomaly detection, we propose \textbf{SwinAD}, a hierarchical reconstruction-based framework that combines multi-scale Transformer representations with feature diversity-preserving reconstruction for robust and accurate anomaly localization. Some visualization examples from MVTec AD (top), VisA (middle), and Real-IAD (bottom) benchmarks along with results from our proposed method are illustrated in Figure \ref{fig:intro_samples}. In summary, the main contributions of this work are as follows:
\begin{itemize}
    \item We propose a hierarchical feature extraction strategy based on a pretrained Swin Transformer V2, which produces multi-scale representations that capture both local textures and global semantics. The encoder is kept frozen to preserve stable and discriminative representations between classes during training.
    \item We introduce a feature diversity-preserving reconstruction framework that alleviates deterministic feature collapse in reconstruction-based anomaly detection. Instead of representing normality using a single reconstruction prototype, the proposed method maintains multiple reconstruction hypotheses to better preserve the diversity of normal patterns, improving robustness to distribution shifts and enhancing discrimination between normal and anomalous local structures.
    \item We evaluate the proposed method on three widely used MUAD benchmarks and demonstrate competitive performance across both image-level and pixel-level metrics. Ablation studies further validate the effectiveness of hierarchical encoding, intermediate feature aggregation, and the feature diversity-preserving reconstruction design.
\end{itemize}

The remainder of this paper is organized as follows: Section \ref{sec:related-works} surveys related works on multi-class unsupervised anomaly detection. Section \ref{sec:methods} presents the proposed method in detail, including the overall framework and its key components. Section \ref{sec:experiments} describes the datasets used in the experiments, the implementation details, evaluation metrics, and the experimental settings. In addition, quantitative and qualitative results are reported and compared with state-of-the-art methods to demonstrate the effectiveness of the proposed approach. Finally, Section \ref{sec:conclusions} concludes the paper and discusses potential directions for future research and further improvements.
\section{Related Works}
\label{sec:related-works}

\subsection{Unsupervised Anomaly Detection}

Visual anomaly detection has been extensively studied for intelligent industrial inspection systems, where the objective is to identify abnormal samples without requiring exhaustive annotations of defect patterns. Existing unsupervised anomaly detection methods can generally be categorized into augmentation-based, embedding-based, and reconstruction-based approaches.

\noindent{\textbf{Augmentation-based Anomaly Detection.}} Augmentation-based methods improve anomaly discrimination by generating synthetic defects during training. These methods expose the model to artificially generated abnormal patterns and formulate anomaly detection as a discrimination problem between normal and synthesized anomalies. CutPaste \cite{cutpaste} generates synthetic defects by randomly cutting image patches and pasting them into different locations within the same image, enabling self-supervised representation learning through anomaly discrimination. DRAEM \cite{draem} synthesizes realistic defect regions and jointly optimizes reconstruction and segmentation networks to improve localization performance. NSA \cite{nsa} synthesizes anomalies in the feature space to generate more diverse abnormal patterns and to enhance generalization to unseen defects. SimpleNet \cite{simplenet} generates anomalous features by perturbing normal embeddings in the feature space and trains a lightweight discriminator to distinguish normal and synthetic anomalous features. By avoiding complex reconstruction or nearest-neighbor retrieval, SimpleNet achieves both strong anomaly detection performance and high inference efficiency. Although these methods effectively improve anomaly discrimination, their performance often depends on the realism and diversity of the generated anomalies.

\noindent{\textbf{Embedding-based Anomaly Detection.}}Embedding-based methods model the distribution of normal samples in a pretrained feature space and identify anomalies through feature discrepancies. PaDiM \cite{padim} models the statistical distribution of patch-level features extracted from pretrained convolutional networks using multivariate Gaussian distributions. During inference, anomaly scores are computed based on the Mahalanobis distance between test features and the learned normal distributions. PatchCore \cite{patchcore} further improves this paradigm by constructing a memory bank of representative normal embeddings and detecting anomalies through nearest-neighbor retrieval in the feature space. 

Building upon feature-space modeling, several studies introduce distillation-based methods, where a student network learns to mimic the representations produced by a pretrained teacher model on normal samples \cite{bergmann2020uninformed, mkd, ast, zhong2025fad}. During inference, anomalies are identified through discrepancies between teacher and student features. STFPM \cite{stpm} aligns multi-scale feature representations between teacher and student networks, enabling anomaly localization through feature inconsistency. Reverse Distillation (RD4AD) \cite{deng2022anomaly} improves this idea by introducing a reconstruction-style student network that recovers teacher representations from compressed embeddings, enhancing anomaly localization. EfficientAD \cite{efficientad} focuses on reducing computational complexity while maintaining competitive anomaly detection performance through lightweight teacher-student architectures and auxiliary training strategies. By leveraging pretrained visual representations, embedding-based methods achieve strong anomaly detection performance and good generalization capability. However, their localization accuracy may be limited when anomalies are subtle or normal patterns are highly diverse. Consequently, recent studies have increasingly focused on reconstruction-based approaches for more accurate anomaly localization.

\noindent{\textbf{Reconstruction-based Anomaly Detection.}} Reconstruction-based methods involve training the model to accurately reconstruct normal images, while producing limited reconstruction capability for defective samples. Early approaches mainly relied on autoencoders \cite{bergmann2018improving, fastrecon, dsr, ma2025patch} and generative adversarial networks (GANs) \cite{kimura2020adversarial, ocgan, stylegan-ad} to reconstruct normal images while producing larger reconstruction errors on anomalous regions. Although effective, image-level reconstruction often struggles to preserve high-level semantic information and may inadvertently reconstruct anomalous regions, reducing anomaly sensitivity.

To address this limitation, recent studies increasingly perform reconstruction in feature space rather than image space, leveraging semantically richer representations extracted from pretrained visual backbones. By reconstructing normal feature distributions, these methods achieve stronger anomaly discrimination and improved localization performance \cite{yang2020dfr, uniad, diad, vitad, he2024mambaad, dinomaly}. DFR \cite{yang2020dfr} reconstructs pretrained CNN features and demonstrates that feature-level reconstruction provides stronger anomaly sensitivity than image-level reconstruction. Recent methods further leverage pretrained transformer representations \cite{dosovitskiy2020image} for feature reconstruction. UniAD \cite{uniad} introduces masked transformer reconstruction for multi-class anomaly detection and demonstrates the effectiveness of transformer-based feature modeling. ViTAD \cite{vitad} utilizes Vision Transformer features to improve representation quality and anomaly localization. More recently, Dinomaly \cite{dinomaly} employs pretrained transformer representations together with noisy bottleneck modules and dropout regularization to mitigate identity mapping and improve reconstruction robustness. MambaAD \cite{he2024mambaad} further explores state-space sequence modeling for efficient feature reconstruction, while diffusion-based approaches such as DiAD \cite{diad} enhance reconstruction quality through iterative denoising processes. Due to their ability to generate dense anomaly maps and provide accurate defect localization, reconstruction-based methods have become one of the most widely adopted paradigms for industrial anomaly detection.

\subsection{Multi-class Anomaly Detection.} 

Most early anomaly detection methods are developed under the Single-class Unsupervised Anomaly Detection (SUAD) setting, where an independent model is trained for each object category. Although such approaches often achieve strong performance, they become increasingly impractical in real-world industrial environments involving numerous product types. Maintaining separate models significantly increases training cost, memory consumption, deployment complexity, and system maintenance overhead. Consequently, recent research has shifted toward Multi-class Unsupervised Anomaly Detection (MUAD) \cite{regad, uninet, lafite, vlmdiff, omnial}, where a single unified model simultaneously learns normal patterns across multiple categories. Compared with SUAD, MUAD introduces substantially greater challenges because normal samples from different categories often exhibit large variations in texture, geometry, appearance, and semantic structure. Therefore, MUAD models must simultaneously preserve category-invariant normality representations while remaining sensitive to subtle anomalous structures.

UniAD \cite{uniad} is one of the earliest transformer-based frameworks designed specifically for MUAD. By introducing neighbor-masked attention and learnable normality queries, UniAD mitigates information leakage during reconstruction and improves generalized anomaly modeling across multiple categories. Building upon this idea, several subsequent studies have explored stronger feature reconstruction strategies and richer pretrained representations. ReContrast \cite{guo2023recontrast}  introduces contrastive reconstruction learning, where contrastive objectives are embedded into feature reconstruction to adapt pretrained representations toward target domains while preventing pattern collapse and identity shortcuts during training. ViTAD \cite{vitad} leverages pretrained Vision Transformer features to improve semantic representation quality, while Dinomaly \cite{dinomaly}  leverages self-supervised DINO-pretrained Vision Transformer representations \cite{caron2021emerging} and introduces noisy bottleneck modules with dropout regularization to prevent identity mapping and enhance anomaly localization. By reconstructing semantically rich DINO features rather than raw images, Dinomaly achieves strong anomaly discrimination and localization performance in multi-class anomaly detection settings. MambaAD \cite{he2024mambaad} explores state-space sequence modeling for efficient feature reconstruction and anomaly detection. Diffusion-based methods \cite{lafite, vlmdiff, diad} improve reconstruction quality through iterative denoising processes and generative modeling. Although recent reconstruction-based methods built upon pretrained Vision Transformer backbone achieve impressive anomaly detection performance, they primarily rely on highly semantic feature representations. While these representations provide strong category-level discrimination and generalization capability, they may suppress fine-grained structural details that are critical for accurately localizing subtle industrial defects. Consequently, achieving a better balance between semantic robustness and localization precision remains an important challenge for industrial anomaly detection.

\section{Methods}
\label{sec:methods}
\begin{figure*}[t]
    \centering
    \includegraphics[width=1.0\linewidth]{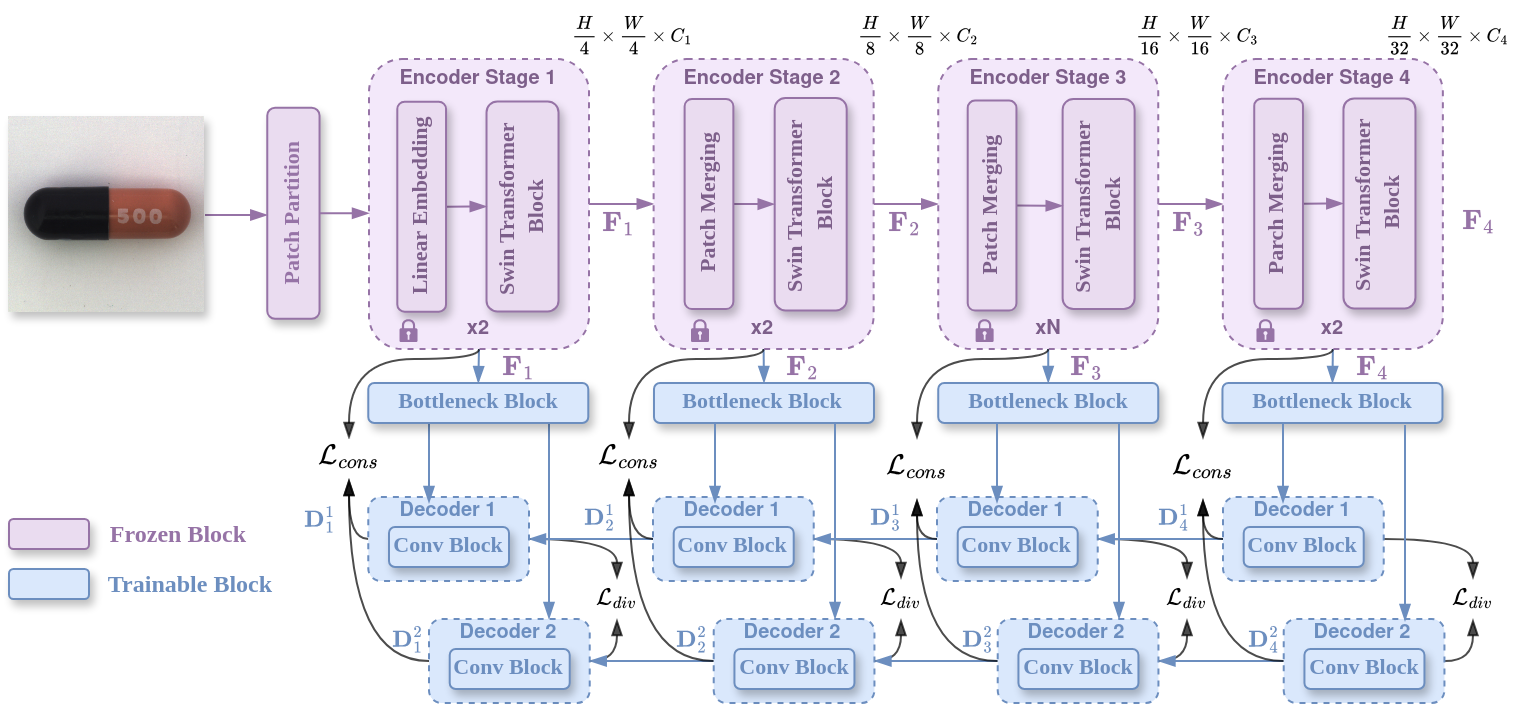}
    \caption{Illustration of the proposed SwinAD framework. A frozen pretrained Swin Transformer V2 first extracts hierarchical multi-scale features from the input image. These features are then aligned by stage-wise bottleneck modules with dropout and passed to a feature diversity-preserving reconstruction decoder, which produces complementary reconstructions at different scales. The differences between the encoder and reconstructed features are finally combined to generate the anomaly map.}
    \label{fig:swinad_architecture}
\end{figure*}
In this section, we present the proposed SwinAD framework for multi-class unsupervised industrial anomaly detection. SwinAD combines hierarchical multi-scale representations extracted from a pretrained Swin Transformer encoder with a feature diversity-preserving reconstruction framework. The hierarchical encoder captures both local structural information and high-level semantic context, while the proposed reconstruction strategy maintains complementary reconstruction hypotheses to better preserve normal feature diversity. As illustrated in Figure \ref{fig:swinad_architecture}, the proposed framework consists of three main components: (1) a hierarchical Swin Transformer encoder for multi-scale feature extraction, and (2) a feature diversity-preserving reconstruction module for robust anomaly localization, and (3) reconstruction and diversity losses that jointly optimize reconstruction fidelity and feature diversity.

\subsection{Problem Formulation}

In this work, we consider the setting of Multi-class Unsupervised Anomaly Detection (MUAD), where the training set contains multiple normal categories but no anomalous samples are available. Formally, given an AD dataset $\mathcal{D}_{train}=\{x_n, y_n\}_{n=1}^{N}$ with class labels $y_n\in \{1,...,C\}$, the normal data distribution is inherently multi-modal. At the test time, images may contain unknown anomalies, and the model must detect them without access to class labels or any supervision about data structure. This setting is particularly challenging because the variation between normal samples of multiple classes could be larger than deviation between anomalies, making it difficult to distinguish unfamiliar normal patterns and truly anomalous ones. Therefore, an effective MUAD model must simultaneously capture class-specific structures while maintaining generalization across all normal categories.

To address this challenge, we leverage a pretrained transformer backbone to extract semantically meaningful features, which already encode general visual structure and implicit category separation. Building upon this representation, the model learns to reconstruct normal patterns without explicit class supervision, and anomalies are detected as deviations from these learned feature distributions. Formally, the goal is to learn a function $f: \mathbf{x} \mapsto \mathcal{A}(\mathbf{x})$ where $\mathcal{A}(\mathbf{x}) \in \mathbb{R}^{H\times W}$ is an anomaly map such that:
\begin{equation*}
    \mathcal{A}(\mathbf{x})\propto \text{Dist}(f_{\text{enc}}(\mathbf{x}), f_{\text{dec}}(\mathbf{x}))
\end{equation*}
where $f_{\text{enc}}$ and $f_{\text{dec}}$ are features from the pretrained backbone and reconstructed features learned by the decoder of the MUAD model, respectively. The key assumption is that normal features are reconstructible, while anomalous patterns typically result in higher reconstruction errors.

\subsection{Swin Transformer as Encoder}

To effectively model the feature distribution described above, the choice of feature representation plays a critical role. Existing reconstruction-based MUAD methods \cite{vitad, dinomaly} often adopt pretrained Vision Transformers (ViT) backbones due to their strong global modeling capability. Although global features effectively capture the high-level semantics and faciliate image-level anomalies, they are less suitable for precise anomaly localization. As a result, these representations are often insufficient for accurately identifying pixel-level anomalies, especially in scenarios involving subtle defects or small abnormal regions.

To address this limitation, we employ the pretrained Swin Transformer V2 model \cite{liu2022swin} as the image encoder. In practice, the encoder is organized into four stages, each stage outputs a hidden representation, resulting in a set of features $[\mathbf{F}_1, \mathbf{F}_2, \mathbf{F}_3, \mathbf{F}_4]$, where $\mathbf{F}_i\in \mathbb{R}^{C_i\times H_i \times W_i}$ with $H_i = \frac{H}{2^i}$ and $W_i = \frac{W}{2^i}$ is the spatial feature map at the \textit{i}-th Swin stage. Since the Swin Transformer V2 model organizes computation hierarchically, the extracted features naturally capture different levels of abstraction, from local textures in shallow layers ($\mathbf{F}_1,\ \mathbf{F}_2$) to semantic structures in deeper layers ($\mathbf{F}_3,\ \mathbf{F}_4$). Importantly, the encoder is kept frozen during training the MUAD model, and no gradients are propagated through its parameters. This design ensures that the pretrained feature space remains stable and preserves the semantic structure learned from large-scale data.

Furthermore, we exploit the hierarchical structure of Swin Transformer V2 model to obtain more robust intermediate representations. In typical configuration, the number of transformer blocks in each stage is [2, 2, N, 2], where the third stage contains significantly more layers than the others. Therefore, Stage 3 forms a deeper transformation sequence, where features are progressively refined through multiple self-attention and feed-forward operations. Using only the final block output biases the representation toward high-level semantics and may suppress fine-grained details. Instead, we aggregate features across all blocks in the third stage by computing $\mathbf{F}_3=\frac{1}{N}\sum_{j=1}^{N} \mathbf{F}_3^j$, where $\mathbf{F}_3^j$ denotes the intermediate feature from the 
\textit{j}-th block in Stage 3.

\subsection{Bottleneck Module}
To bridge the representation gap between the encoder and decoder features, we apply a stage-wise bottleneck transformation. Formally, given an encoder feature $\mathbf{F}_i$, the aligned feature is computed as $\mathbf{Z}_i=\mathcal{B}_i(\mathbf{F}_i)$, where $\mathcal{B}_i(\cdot)$ is a learnable bottleneck block with dropout regularization. The design of the bottleneck module is directly adopted from \cite{dinomaly} without significant modifications. 

The use of dropout is particularly important in unsupervised anomaly detection; without it, the model tends to learn a near-identity mapping ($\mathbf{D}_i\approx \mathbf{F}_i$, where $\mathbf{D}_i$ is corresponding decoder feature). Therefore, the decoder simply copies the encoder features. In this case, both normal and anomalous samples could be reconstructed equally well, leading to low reconstruction error for anomalies and thus poor anomaly discrimination. By employing dropout, part of the feature information is randomly suppressed during training, forcing the decoder to learn a more robust and generalized reconstruction of normal patterns rather than relying on exact feature matching. As a result, normal samples could be reconstructed accurately, while anomalies produce higher reconstruction error during the inference phase, improving the performance of anomaly detection.

\subsection{Feature Diversity-Preserving Reconstruction}

Existing reconstruction-based anomaly detection methods assume that normal features can be reconstructed accurately while anomalous features cannot. Given an encoder feature map $\mathbf{F}$, a decoder reconstructs a normal feature projection $\mathbf{D}$, and an anomaly map $\mathbf{A}$ is estimated using pixel-wise feature discrepancy:
\begin{equation*}
    \mathbf{A}_{ij} = d(\mathbf{F}_{ij}, \mathbf{D}_{ij}) = 1 - \text{cos}(\mathbf{F}_{ij}, \mathbf{D}_{ij})
\end{equation*}
where $\mathbf{F}_{ij}$ and $\mathbf{D}_{ij}$ denote the encoder and reconstructed features at spatial location $(i, j)$. Under this formulation, anomaly detection becomes a point-wise compatibility problem and the model implicitly assumes that each local normal structure can be represented by one deterministic reconstruction projection. However, local normal structures in industrial images are inherently diverse. For example, even within the same object category, normal regions may contain different textures, materials, illumination conditions, and local geometric patterns. Therefore, the local normal distribution forms a manifold with multiple valid modes rather than a single compact representation:
\begin{equation*}
    p(x_{ij} | z_{ij}) = \sum_k \pi_k p_k(x_{ij}|z_{ij})
\end{equation*}

where each component $p_k(x_{ij}|z_{ij})$ represents a different valid local normal mode. Under single reconstruction learning, the decoder gradually compresses these multiple normal modes into one dominant semantic projection $\mathcal{M}_{normal} \rightarrow D$. This behavior becomes problematic when the testing distribution differs from the training distribution, since unseen but still normal local structures may deviate from the collapsed reconstruction prototype despite remaining semantically normal. We argue that the major limitation of existing reconstruction-based methods is not insufficient reconstruction capability, but excessive representation collapse caused by deterministic reconstruction learning. Consequently, the model loses tolerance to normal structural variations and becomes less sensitive to subtle local inconsistencies. 

To address this issue, the proposed method preserves normal feature diversity using multiple manifold-consistent reconstruction hypotheses $\mathbf{D}^1_{ij}$ and $\mathbf{D}^2_{ij}$, inspired by recent findings in representation learning that complementary views can mitigate representation collapse and improve generalization \cite{zhu2021class, harun2025controlling}. Instead of collapsing the normal manifold into one deterministic projection $\mathcal{M}_{normal} \rightarrow \mathbf{D}$, the proposed formulation preserves a broader manifold structure:
\begin{equation*}
    \mathcal{M}_{normal} \rightarrow (\mathbf{D}^1, \mathbf{D}^2)
\end{equation*}
Importantly, the objective is not simply to generate multiple reconstructions, but to preserve feature diversity and manifold coverage during reconstruction learning. Each reconstruction hypothesis captures a different valid local normal structure while remaining consistent with the underlying normal manifold:
\begin{equation*}
    \mathbf{D}^1,\  \mathbf{D}^2 \in \mathcal{M}_{normal},\mathbf{D}^1\neq \mathbf{D}^2
\end{equation*}

This diversity substantially improves generalization capability because unseen but still normal local features may remain compatible with at least one manifold-consistent projection. In contrast, anomalous local structures do not belong to the learned normal manifold. Although anomaly features may still preserve partial similarity to one projection individually due to dominant normal semantics, they become significantly less consistent with the manifold geometry jointly represented by multiple hypotheses. Therefore, anomaly estimation is naturally transformed from point-wise similarity estimation into manifold compatibility estimation:
\begin{equation*}
    \mathbf{A}_{ij} = d(\mathbf{F}_{ij}, \mathcal{M}(D^1_{ij},D^2_{ij}) = 1-\text{cos}(\mathbf{F}_{ij} \oplus \mathbf{F}_{ij}, \mathbf{D}^1_{ij} \oplus \mathbf{D}^2_{ij})
\end{equation*}

where $\oplus$ indicates the concatenation operator. By maintaining feature diversity during reconstruction learning, the model becomes substantially more robust to distribution shifts while simultaneously improving discrimination between normal and anomalous local structures. In this paper, we employ two reconstruction hypotheses to provide a practical balance between feature diversity and optimization stability. The objective of the proposed method is not to fully model the entire normal distribution using many decoders, but to prevent deterministic feature collapse caused by single reconstruction learning. Moreover, increasing the number of decoder branches significantly increases computational complexity, memory consumption, and optimization difficulty during training and inference. Therefore, using only two reconstruction hypotheses provides an effective trade-off between manifold diversity, generalization capability, anomaly discrimination, and computational efficiency. In our implementation, we directly adopt the lightweight decoder block from \cite{dinomaly}. Features from deeper layers are progressively upsampled and passed through a $1\times1$ convolution layer before being fused with shallower encoder features.

\subsection{Loss function}
To preserve local manifold diversity while improving sensitivity to subtle structural inconsistencies, we employ a hard-mining local cosine similarity loss for reconstruction learning. Given the encoder feature map $\mathcal{F}_s \in \mathbb{R}^{H_s\times W_s \times C_s}$ at scale $s$, the two decoder branches independently reconstruct manifold-consistent feature representations $\mathbf{D}^1_s$ and $\mathbf{D}^2_s$. Instead of jointly optimizing both reconstructions, we compute the reconstruction discrepancy separately for each decoder branch in order to encourage each decoder to preserve different valid local manifold structures. 

Motivated by previous work, we focus the optimization on difficult local regions with larger reconstruction discrepancy. Specifically, for each decoder branch $m \in \{1,2\}$ and feature at stage $s\in\{1,2,3,4\}$, we select a set of top-k hardest spatial tokens $\Omega^m_{s,k}$. The reconstruction loss for each decoder branch is then formulated as:
\begin{equation*}
    \mathcal{L}^m_s = \frac{1}{|\Omega^m_{s,k}|}\sum_{(i,j)\in \Omega^m_{s,k}} (1 - \text{cos}(\mathbf{F}_{s,ij}, \mathbf{D}^m_{s,ij}))
\end{equation*}

The multi-scale reconstruction objective is defined as:
\begin{equation*}
    \mathcal{L}_{rec} = \sum_{s=1}^4(\mathcal{L}_s^1 + \mathcal{L}_s^2)
\end{equation*}

To preserve manifold diversity between the two reconstruction branches, we introduce a cosine-based divergence regularization that explicitly discourages the reconstructed features from collapsing into identical representations. For each feature scale $s$, the diversity loss is formulated as:
\begin{equation*}
    \mathcal{L}_{\text{div}} = \text{max}(0,\ \text{cos}(\mathbf{D}^1_s, \mathbf{D}^2_s) - \tau_s)^2
\end{equation*}
where $\tau_s$ is a stage-dependent margin. This prevents the two features from collapsing into identical representations and promotes complementary modeling of normal patterns. In our implementation, we empirically set $\tau_s \in \{0.4, 0.8, 0.95, 0.99\}$ for different feature scales from shallow to deep layers. Lower-level features typically contain richer local textures and structural variations, requiring stronger diversity constraints, while deeper semantic features are naturally more compact and therefore use larger similarity margins. We further incorporate the divergence loss with a progressively increasing weight: 
\begin{equation*}
    \mathcal{L}_{\text{final}} = \mathcal{L}_{\text{rec}} + \lambda_{\text{div}}\mathcal{L}_{\text{div}},\ \lambda_{\text{div}} = \text{min}(\lambda \cdot \frac{t}{1000},\ \lambda) 
\end{equation*}
where $t$ is the training iteration. This progressive scheduling strategy stabilizes optimization during early training stages by allowing the model to first learn stable reconstruction of normal patterns before encouraging diversity between reconstructed features. Early in training, a small divergence weight allows the model to learn consistent normal representations. As training progresses, the increasing weight encourages the decoder branches to capture complementary features without collapsing, leading to more meaningful diversity and improved anomaly discrimination.

\subsection{Anomaly Map Estimation}
\begin{figure}[pos=tbp]
    \centering
    \includegraphics[width=0.9\linewidth]{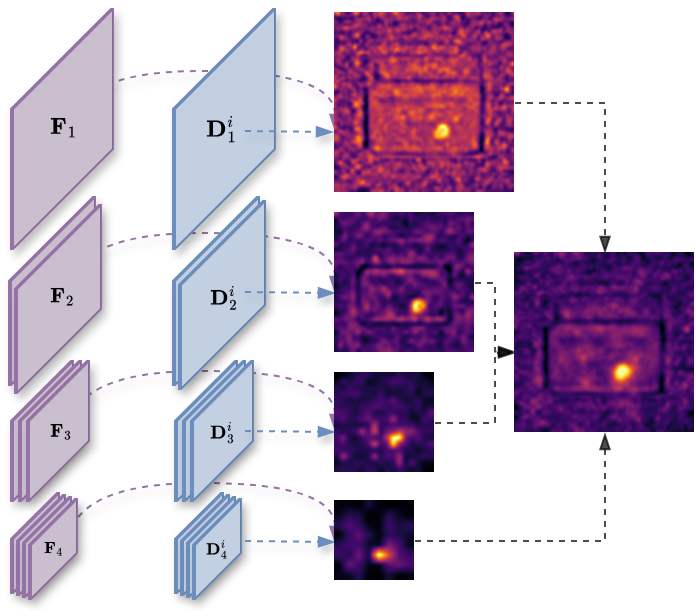}
    \caption{Overview of the anomaly map estimation process of SwinAD. For each stage, the encoder feature is compared with the two reconstructed features from the dual decoder using cosine distance. The scale-wise anomaly maps are first resized to a unified resolution and then aggregated into a single map.}
    \label{fig:anomaly_map_extraction}
\end{figure}
To estimate the anomaly map, we measure the discrepancy between reconstructed features and their corresponding target features across multiple scales. Let $\mathbf{F}_i$ and $\mathbf{D}_s^1$, $\mathbf{D}_s^2$ ($s \in \{1,..,N\}$) denote the encoder and two reconstructed feature maps from two decoder branches at the $s$-th level, respectively. For each level, we compute a pixel-wise anomaly response using cosine distance:
\begin{equation*}
    \mathbf{A}_{s,ij} = 1- \text{cos}((\mathbf{F}_{s,ij} \oplus \mathbf{F}_{s,ij}), (\mathbf{D}^1_{s,ij} \oplus \mathbf{D}_{s,ij}^2))
\end{equation*}

The resulting map $\mathbf{A}_s \in \mathbb{R}^{H_s\times W_s}$ is then resized to a target resolution $(H_t, W_t)$ using bilinear interpolation. Each scale is further weighted by a predefined factor $w_i$ ($\sum_{i=1}^N w_i = N$), and the final anomaly map is obtained by aggregating anomaly maps at different levels:
\begin{equation*}
    \mathbf{A} = \frac{1}{N} \sum_{i=1}^N w_i \cdot \text{Interp}(\mathbf{A}_i, H_t, W_t)
\end{equation*}
Figure \ref{fig:anomaly_map_extraction} illustrates the multi-scale anomaly map estimation process of the proposed framework. Shallow feature levels mainly capture low-level visual patterns such as edges, textures, local intensity variations, and fine structural details. Therefore, anomaly maps generated from early feature stages usually exhibit higher spatial resolution and stronger sensitivity to subtle local defects. However, because shallow features primarily focus on local appearance information, their anomaly responses are often noisier and more sensitive to texture variations or illumination changes. As shown in Figure \ref{fig:anomaly_map_extraction}, shallow-level anomaly maps preserve fine-grained boundary information and highlight detailed local irregularities around the defective region. In contrast, deeper feature levels encode more abstract semantic representations and larger contextual structures. Instead of focusing on local texture patterns, deeper layers capture global object consistency, semantic layouts, and high-level structural relationships. Therefore, anomaly maps generated from deeper stages usually produce more compact and semantically stable responses. Although their spatial resolution becomes coarser due to progressive feature downsampling, deeper anomaly maps tend to suppress irrelevant local noise and emphasize semantically inconsistent regions more clearly. By aggregating anomaly evidence across multiple representation levels, the proposed framework simultaneously preserves fine local details and high-level semantic consistency, resulting in sharper, more robust, and spatially coherent anomaly localization.
Finally, to enhance spatial consistency and suppress noise, a Gaussian smoothing operation $\mathcal{G}(\cdot)$ is applied:
\begin{equation*}
    \mathbf{A}_{\text{final}} = \mathcal{G}(\mathbf{A})
\end{equation*}
In summary, this formulation enables multi-scale anomaly detection by combining feature discrepancies across different representation levels while ensuring spatial coherence in the final prediction.

\section{Experiments}
\label{sec:experiments}
\subsection{Dataset}
We conduct experiments on three widely used industrial anomaly detection datasets, including MVTecAD \cite{mvtecad}, VisA \cite{visa}, and Real-IAD \cite{wang2024real}, supporting both image-level classification and pixel-level segmentation tasks. 

\noindent{\textbf{MVTecAD Dataset}} \cite{mvtecad}: The MVTec AD (MVTec Anomaly Detection) dataset consists of 15 categories spanning object-based classes (eg., bottle, capsule, metal nut) and texture-based classes (e.g., carpet, grid, leather), each containing diverse defect types such as broken, scratch, crack, and contamination, etc. MVTec AD follows an unsupervised learning setting, where only normal samples are available during training, while the test set contains both normal and anomalous images with pixel-level ground-truth annotations. Due to its diverse defect patterns and high-quality annotations, MVTec AD has become one of the most widely used benchmarks for industrial anomaly detection.

\noindent{\textbf{VisA Dataset}} \cite{visa}: The VisA (Visual Anomaly) dataset is designed to reflect more realistic and diverse industrial scenarios. It contains 12 object categories with substantial variations in illumination, background clutter, and viewpoint. Similar to MVTec AD, VisA adopts an unsupervised training protocol using only normal images, while providing pixel-level annotations for both normal and anomalous samples during evaluation. Compared with MVTec AD, VisA significantly increases the difficulty by incorporating more complex scenes and subtle anomalies that may depend on contextual or semantic cues rather than purely local texture differences. As a result, it serves as a stronger benchmark for assessing the robustness and generalization capability of anomaly detection models in real-world scenarios.

\noindent{\textbf{Real-IAD Dataset}} \cite{wang2024real}: The Real-IAD (Real-world Industrial Anomaly Detection) dataset is a large-scale real-world industrial anomaly detection benchmark with 30 classes designed to better reflect practical manufacturing environments. Compared with existing datasets such as MVTec AD and VisA, Real-IAD contains significantly more complex industrial scenarios with larger intra-class variations, cluttered backgrounds, illumination changes, viewpoint diversity, and subtle defect patterns. In particular, anomalies in Real-IAD are often small, sparse, and highly similar to surrounding normal structures, making anomaly localization substantially more challenging. These characteristics increase the difficulty of distinguishing anomalous regions from diverse normal patterns. Therefore, Real-IAD provides a more realistic and challenging benchmark for evaluating the robustness, generalization capability, and localization performance of industrial anomaly detection methods under real-world distribution variations.

All three datasets provide image-level labels together with pixel-level anomaly ground-truth masks, enabling evaluation on both image-level anomaly classification and pixel-level anomaly localization tasks, providing comprehensive evaluation protocols for industrial anomaly detection.

\subsection{Evaluation Metrics}
Following previous works \cite{zhang2024ader}, we employ both threshold-independent and threshold-dependent metrics to evaluate our proposed SwinAD and state-of-the-art methods. We also use both image-level and pixel-level metrics to provide a comprehensive evaluation. At the image level, we adopt three metrics, including the mean Area Under the Receiver Operating Characteristic (I-AUROC), mean Average Precision (I-AP), and mean F1 Score (I-F1) to measure the model's ability to distinguish between normal and anomalous images. Specifically, I-AUROC evaluates overall discrimination capability across different thresholds, I-AP summarizes the precision-recall trade-off, and I-F1 reflects the balance between precision and recall at the optimal threshold. At the pixel level, we employ similar metrics (P-AUROC, P-AP, and P-F1) to comprehensively evaluate anomaly localization performance. While P-AUROC and P-AP quantify pixel-wise discrimination and precision-recall characteristics, P-F1 measures segmentation performance at the optimal threshold. 

Furthermore, we focus mainly on pixel-level metrics for evaluation. Although image-level metrics can indicate whether a model classifies an image as normal or anomalous, they do not reflect whether the detected anomalous regions are spatially accurate. In practice, a model may correctly label an image as anomalous while highlighting incorrect or irrelevant regions. Therefore, pixel-level metrics provide a more informative assessment of localization performance, as they directly evaluate how accurately the model localizes anomalies, which is especially important for applications that require precise defect detection. In this work, we compare the results of our proposed SwinAD with recent state-of-the-art reconstruction-based multi-class unsupervised anomaly detection methods, including UniAD \cite{uniad}, ReContrast \cite{guo2023recontrast}, DiAD \cite{diad}, ViTAD \cite{vitad}, MambaAD \cite{he2024mambaad}, and Dinomaly \cite{dinomaly}. Unless otherwise specified, the quantitative results of competing methods are directly adopted from their original publications. To provide a fairer and more comprehensive analysis of computational efficiency and localization performance, we additionally reproduce Dinomaly under a 224$\times$224 input resolution setting.

\subsection{Implementation Settings}
The proposed SwinAD framework is implemented using PyTorch and trained on a single NVIDIA RTX A5000 GPU. Following the standard unsupervised anomaly detection protocol, only normal samples are used during training, while both normal and anomalous samples are used for evaluation. All input images are resized to $256 \times 256$ before being fed into the network. No additional data augmentation is applied during training to ensure a fair comparison with existing MUAD methods. For feature extraction, we employ the Swin Transformer V2-Base model pretrained on ImageNet \cite{russakovsky2015imagenet} as the encoder backbone. For reconstruction learning, the top-k hard mining ratio is set to $k=10\%$ of the spatial tokens at each feature scale. For anomaly map estimation, the feature-level aggregation weights are empirically set by $w_i \in \{0.75,\ 0.75,\ 1.25,\ 0.25\}$. The final anomaly map is further refined using Gaussian smoothing with a kernel size of 9$\times$9 and $\sigma=2$.

The bottleneck and decoder modules are optimized using the AdamW \cite{loshchilov2017decoupled} optimizer with an initial learning rate of $1\times 10^{-3}$. The batch size is set to 16, and the network is trained for 100 epochs. A cosine annealing learning rate scheduler is employed to stabilize optimization during training. 

For evaluation, we report both image-level and pixel-level anomaly detection performance using AUROC, Average Precision (AP), and $F_1$-score. The optimal dataset-level threshold is selected for $F
_1$ computation following common evaluation protocols in industrial anomaly detection.

\begin{table*}[pos=tbp]
\centering
\caption{Quantitative comparison on MVTec AD under the multi-class unsupervised anomaly detection setting. We report image-level AUROC, AP, and $F_1$, together with pixel-level AUROC, AP, and $F_1$. The best results are shown in bold, and the values in parentheses indicate the performance difference between SwinAD and the strongest Dinomaly setting.}
\setlength{\tabcolsep}{4pt} 
\renewcommand{\arraystretch}{0.95} 
\small
\begin{tabular}{p{0.15\textwidth}
      >{\centering}p{0.1\textwidth}
      >{\centering}p{0.1\textwidth}
      >{\centering}p{0.1\textwidth}
      >{\centering}p{0.1\textwidth}
      >{\centering}p{0.1\textwidth}
      >{\centering}p{0.1\textwidth}
      >{\centering\arraybackslash}p{0.1\textwidth}
      }
\toprule
\multirow{2}{*}{Method} & \multirow{2}{*}{Image Size}
& \multicolumn{3}{c}{Image-level} 
& \multicolumn{3}{c}{Pixel-level} \\
\cmidrule(lr){3-5} \cmidrule(lr){6-8}
& & AUROC & AP & $F_1$ 
& AUROC & AP & $F_1$ \\
\midrule

UniAD~\cite{uniad}  & $256 \times 256$      & 96.5 & 98.8 & 96.2 & 96.8 & 43.4 & 49.5 \\
ReContrast \cite{guo2023recontrast} & $256 \times 256$ & 98.3 & 99.4 & 97.6 & 97.1 & 60.2 & 61.5 \\
DiAD~\cite{diad}  & $256 \times 256$        & 97.2 & 99.0 & 96.5 & 96.8 & 52.6 & 55.5 \\
ViTAD~\cite{vitad}    & $256 \times 256$    & 98.3 & 99.4 & 97.3 & 97.7 & 55.3 & 58.7 \\
MambaAD \cite{he2024mambaad} & $256 \times 256$   & 98.6 & 99.6 & 97.8 & 97.7 & 56.3 & 59.2\\
Dinomaly \cite{dinomaly} & $224 \times 224$ & 99.4 & 99.8 & 99.0 & 98.1 & 63.0 & 64.7 \\

Dinomaly \cite{dinomaly} & $392 \times 392$  & \textbf{99.6} & \textbf{99.8} & \textbf{99.0} 
                          & \textbf{98.4} & 69.3 & 69.2 \\
\midrule
\textbf{SwinAD (ours)} & $256 \times 256$ & 99.4 \footnotesize (-\textbf{\textcolor{red}{0.2}}) & \textbf{99.8} & 98.8 \footnotesize (-\textbf{\textcolor{red}{0.2}})  & 98.1 \footnotesize (-\textbf{\textcolor{red}{0.3}}) & \textbf{74.4} \footnotesize (+\textbf{\textcolor{brightgreen}{5.1}}) & \textbf{70.9} (+  \footnotesize \textbf{\textcolor{brightgreen}{1.7}})  \\
\bottomrule

\end{tabular}
\vspace{-2mm}
\label{tab:mvtec_results}
\end{table*}

\begin{figure}[pos=tbp]
    \centering
    \includegraphics[width=1.0\linewidth]{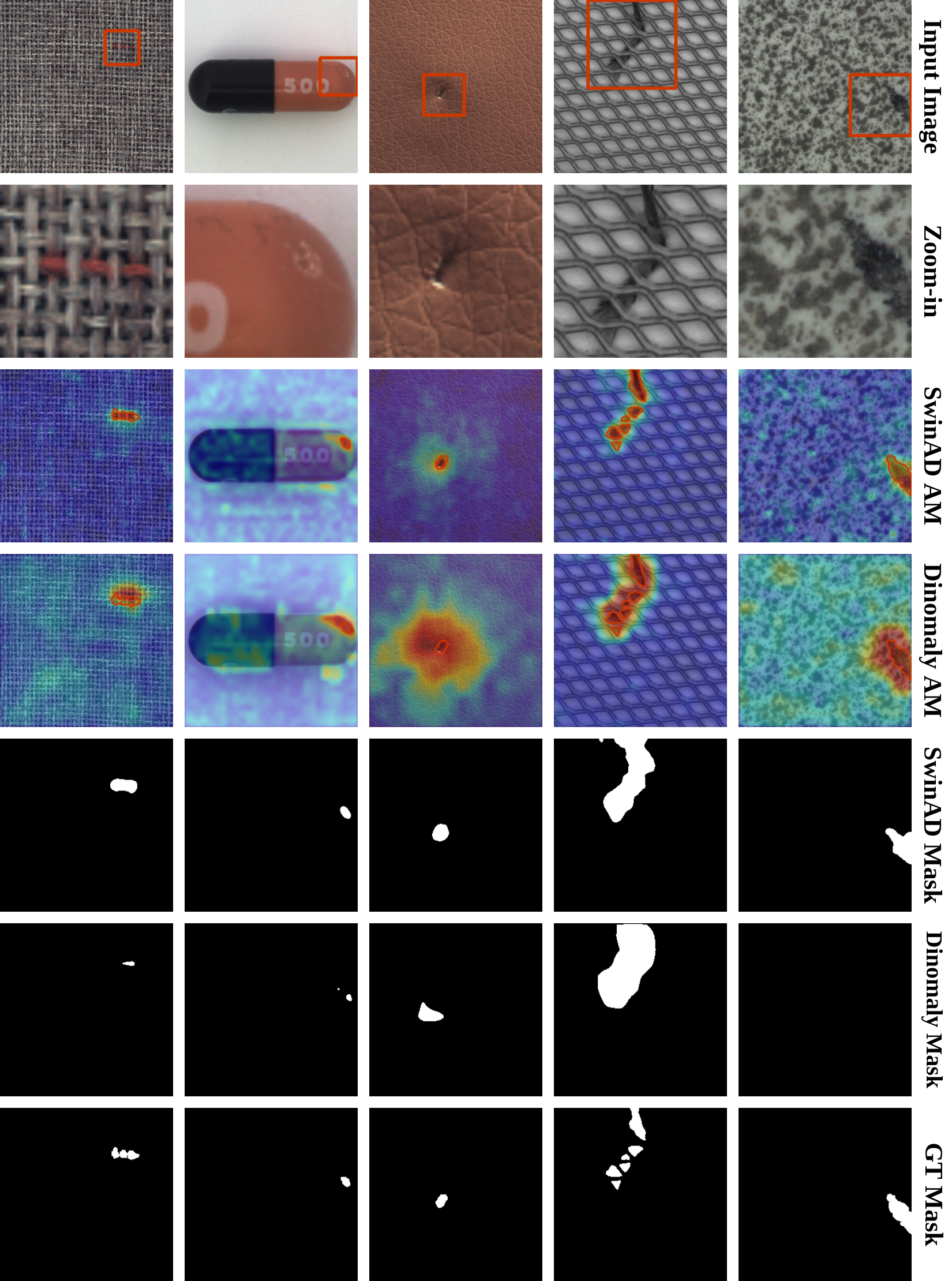}
    \caption{Qualitative comparison of anomaly localization on MVTecAD samples. From top to bottom: input images with annotated defect regions, zoomed-in views of anomalies, anomaly maps (AMs) produced by our proposed SwinAD and Dinomaly, corresponding binary masks after thresholding, and ground-truth (GT) masks. SwinAD generates more compact and precise anomaly responses with fewer background activations, whereas Dinomaly often produces over-smoothed and spatially diffuse predictions, particularly on textured surfaces and elongated defects.}
    \label{fig:mvtec_qualitative}
\end{figure}

\subsection{Experimental Results}
\noindent{\textbf{Evaluations on MVTec AD}}: Table \ref{tab:mvtec_results} presents a comprehensive comparison between our method and state-of-the-art multi-class unsupervised anomaly detection approaches on both image-level and pixel-level anomaly detection metrics on MVTec AD dataset.

Overall, SwinAD demonstrates  competitive performance across all metrics. At the image level, our method achieves an AUROC of 99.4\%, an AP of 99.7\%, and an F1 score of 98.8\%, indicating that the proposed framework can effectively distinguish anomalous and normal samples at the global level.  Although Dinomaly achieves slightly better performance on several image-level metrics, the performance gap remains relatively small, demonstrating that the proposed method maintains strong global anomaly discrimination ability.

More importantly for industrial inspection tasks, SwinAD demonstrates clear advantages in pixel-level anomaly localization. Specifically, the proposed method achieves a pixel-level AP of 74.4\%, which is the highest among all compared methods. Compared with previous approaches, SwinAD improves pixel-level AP by \textbf{\textcolor{brightgreen}{5.1}}\% over Dinomaly, \textbf{\textcolor{brightgreen}{14.2}}\% over ReContrast, and \textbf{\textcolor{brightgreen}{18.1}}\% over MambaAD. Similar improvements can also be observed in the pixel-level $F_1$ metric, further confirming the effectiveness of the proposed framework in generating accurate and spatially consistent anomaly maps. These improvements are particularly important for industrial inspection systems, where precise localization of defective regions is often more critical than image-level classification alone. In many practical applications, incorrect localization may lead to inaccurate defect analysis, unnecessary product rejection, or failure to identify subtle defects. Another important observation is that Dinomaly performs inference using a higher input resolution of $392 \times 392$, while SwinAD operates at a lower resolution of $256 \times 256$. Despite this lower input resolution, the proposed method still achieves comparable image-level performance and superior pixel-level localization accuracy. These results demonstrate that the proposed framework provides an effective trade-off between computational efficiency and anomaly detection performance. The strong performance at a lower resolution also suggests that the hierarchical Swin Transformer representations are capable of preserving rich semantic and structural information even under reduced computational complexity. Furthermore, compared with transformer-based baselines such as ViTAD and MambaAD, our method consistently improves pixel-level metrics by large margins, especially in AP and $F_1$-score, demonstrating the advantage of our architecture in capturing local anomaly structures.

To further analyze the behavior of the proposed framework, we conduct qualitative comparisons mainly against Dinomaly for clearer analysis, since Dinomaly significantly outperforms earlier methods on both image-level and pixel-level anomaly detection metrics. This comparison allows us to focus on the differences in anomaly localization behavior between methods. By analyzing the anomaly maps shown in Figure \ref{fig:mvtec_qualitative}, it can be observed that the anomaly map generated by Dinomaly tends to activate not only on the actual defect regions but also on surrounding normal areas, resulting in redundant and spatially over-expanded responses. In several examples, the anomalous regions are highlighted with excessively large and diffuse activation patterns, even when the true defects occupy only a small localized area. This behavior increases the overall anomaly intensity of the image and makes anomalous samples easier to distinguish from normal samples at the image level. Consequently, Dinomaly achieves strong image-level performance since the aggregated anomaly score remains sufficiently high to correctly classify defective images.

\begin{table*}[t!]
\centering
\caption{Quantitative comparison on VisA dataset. We evaluate both image-level anomaly detection and pixel-level segmentation metrics. SwinAD obtains competitive image-level performance and improves pixel-level AP and $F_1$, demonstrating stronger localization precision at $256\times256$ input resolution.}
\setlength{\tabcolsep}{4pt} 
\renewcommand{\arraystretch}{0.95} 
\small
\begin{tabular}{p{0.15\textwidth}
      >{\centering}p{0.1\textwidth}
      >{\centering}p{0.1\textwidth}
      >{\centering}p{0.1\textwidth}
      >{\centering}p{0.1\textwidth}
      >{\centering}p{0.1\textwidth}
      >{\centering}p{0.1\textwidth}
      >{\centering\arraybackslash}p{0.1\textwidth}
      }
\toprule
\multirow{2}{*}{Method} & \multirow{2}{*}{Image Size}
& \multicolumn{3}{c}{Image-level} 
& \multicolumn{3}{c}{Pixel-level} \\
\cmidrule(lr){3-5} \cmidrule(lr){6-8}
& & AUROC & AP & $F_1$ 
& AUROC & AP & $F_1$ \\
\midrule
UniAD~\cite{uniad}  & $256 \times 256$      & 88.8 & 90.8 & 85.8 & 98.3 & 33.7 & 39.0 \\
ReContrast \cite{guo2023recontrast} & $256 \times 256$ & 95.5 & 96.4 & 92.0 & 98.5 & 47.9 & 50.6 \\
DiAD~\cite{diad}  & $256 \times 256$        & 86.8 & 88.3 & 85.1 & 96.0 & 26.1 & 33.0 \\
ViTAD~\cite{vitad}    & $256 \times 256$    & 90.5 & 91.7 & 86.3 & 98.2 & 36.6 & 41.1 \\
MambaAD \cite{he2024mambaad} & $256 \times 256$   & 94.3 & 94.5 & 89.4 & 98.5 & 39.4 & 44.0 \\

Dinomaly \cite{dinomaly} & $224 \times 224$ & 96.9 & 97.3 & 93.5 & 98.4 & 45.6 & 49.0 \\

Dinomaly \cite{dinomaly} & $392 \times 392$  & \textbf{98.7} & \textbf{98.9} & \textbf{96.2} 
                          & \textbf{98.7} & 53.2 & 55.7 \\
\midrule
\textbf{SwinAD (ours)} & $256 \times 256$ & 
98.0 \footnotesize (-\textbf{\textcolor{red}{0.7}}) & 
98.2 \footnotesize (-\textbf{\textcolor{red}{0.7}}) & 
95.3 \footnotesize (-\textbf{\textcolor{red}{0.9}})  & 
97.0 \footnotesize (-\textbf{\textcolor{red}{1.7}}) & 
\textbf{58.0} \footnotesize (+\textbf{\textcolor{brightgreen}{4.8}}) & \textbf{57.3} (+  \footnotesize \textbf{\textcolor{brightgreen}{1.6}})  \\
\bottomrule

\end{tabular}
\vspace{-2mm}
\label{tab:visa_results}
\end{table*}

\begin{figure}[pos=!htbp]
    \centering
    \includegraphics[width=1.0\linewidth]{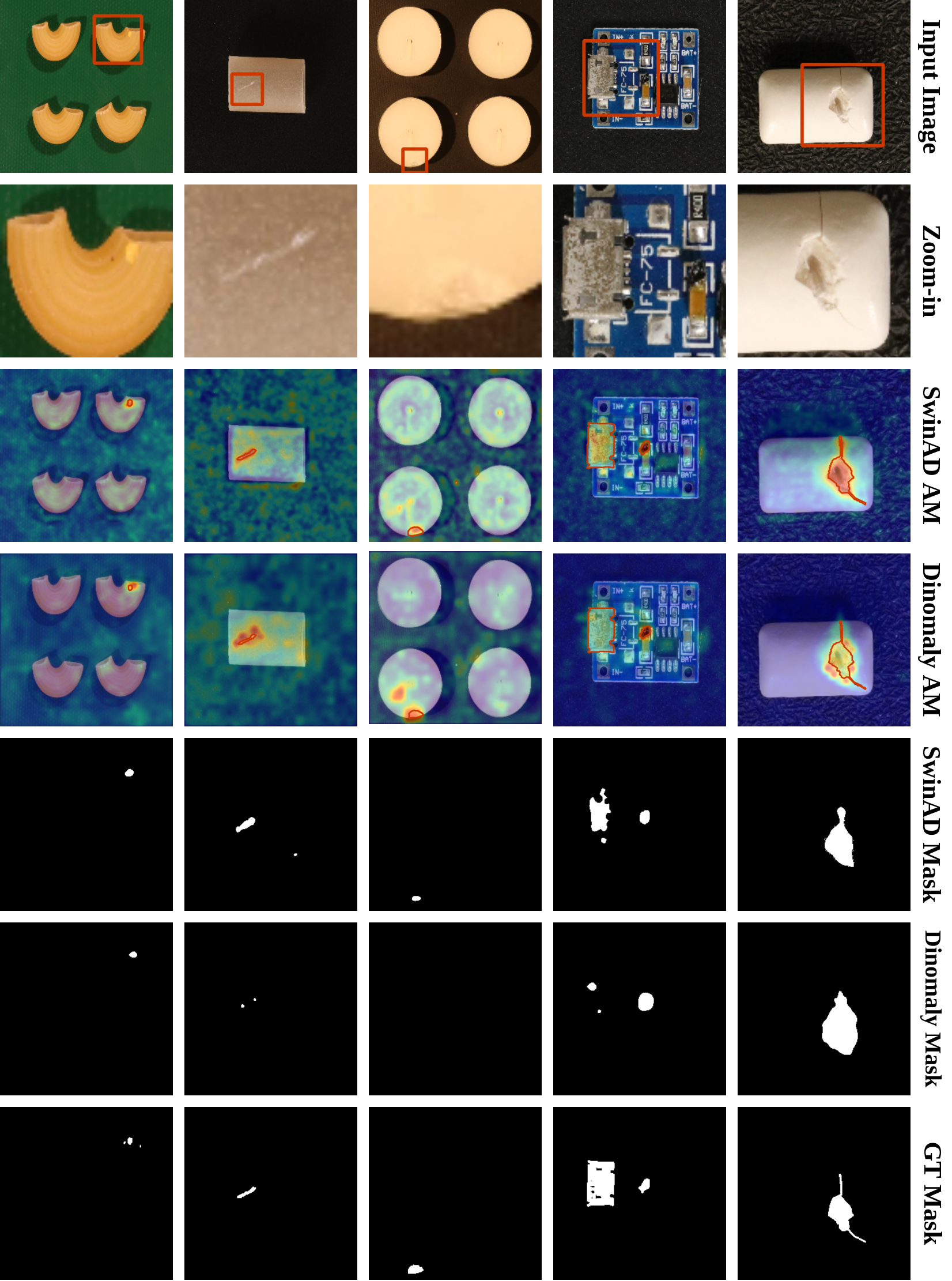}
    \caption{Qualitative comparison of anomaly localization on VisA samples. The rows show input images with annotated defect regions, zoomed-in anomaly regions, anomaly maps produced by SwinAD and Dinomaly, thresholded binary masks, and ground-truth masks. SwinAD produces more localized and compact responses under complex backgrounds and object variations, while Dinomaly tends to generate broader activations around normal regions.}
    \label{fig:visa_qualitative}
\end{figure}

However, these redundant activations negatively affect pixel-level localization accuracy. Because the predicted anomalous regions extend beyond the ground-truth boundaries, the method introduces a considerable number of false positive pixels and produces less precise segmentation masks. As a result, although Dinomaly performs well on image-level metrics, its pixel-level metrics are relatively lower due to inaccurate boundary localization and reduced spatial consistency. Furthermore, pixel-level anomaly detection is inherently highly imbalanced because anomalous pixels typically occupy only a small portion of the image. Under such imbalance, P-AUROC may remain high even when anomaly maps contain spatially diffuse activations, since most anomalous pixels still receive higher scores than normal pixels. In contrast, AP and $F_1$-score are more sensitive to false-positive predictions and localization precision. Therefore, the higher P-AP and P-F1-scores of the proposed method suggest more accurate and spatially consistent anomaly localization despite slightly lower P-AUROC values.  As demonstrated by the generated segmentation masks under the optimal dataset-level threshold and their alignment with the ground-truth annotations, the proposed framework produces more precise and spatially consistent anomaly localization.

\noindent{\textbf{Evaluations on VisA}}: Table \ref{tab:visa_results} presents the quantitative comparison on the VisA dataset. Compared with MVTec AD, VisA contains more challenging industrial scenarios with larger viewpoint variations, cluttered backgrounds, illumination changes, and more subtle anomaly patterns. Furthermore, anomalous regions in the VisA dataset generally occupy a smaller portion of the image and are often more subtle and difficult to localize. These characteristics make accurate anomaly localization substantially more difficult, especially for methods that rely heavily on local texture reconstruction or over-generalized feature representations. 

Overall, the proposed SwinAD framework achieves highly competitive performance across both image-level and pixel-level evaluation metrics. At the image level, SwinAD achieves an AUROC of 98.0\%, an AP of 98.2\%, and an $F_1$-score of 95.3\%. Although Dinomaly operating at a higher input resolution of $392 \times 392$ achieves  better image-level performance, the performance gap remains relatively small despite SwinAD using a lower input resolution of $256 \times 256$. These results demonstrate that the proposed hierarchical reconstruction framework maintains strong anomaly discrimination capability even under more complex visual conditions and distribution variations. Under the highly imbalanced setting between normal and anomalous samples, Dinomaly achieves stronger image-level performance because its broader anomaly activations may increase the global separability between anomalous and normal samples

More importantly, SwinAD demonstrates clear advantages in pixel-level anomaly localization. The proposed method achieves the best pixel-level AP of 58.0\% and the highest pixel-level $F_1$-score of 57.3\%, outperforming Dinomaly by \textbf{\textcolor{brightgreen}{4.8}}\% and \textbf{\textcolor{brightgreen}{1.6}}\%, respectively. Significant improvements are also observed compared with other recent methods such as ReContrast (+\textbf{\textcolor{brightgreen}{10.1}}\% P-AP, +\textbf{\textcolor{brightgreen}{6.7}}\% P-$F_1$), MambaAD (+\textbf{\textcolor{brightgreen}{18.6}}\% P-AP, +\textbf{\textcolor{brightgreen}{13.3}}\% P-$F_1$), and ViTAD (+\textbf{\textcolor{brightgreen}{21.4}}\% P-AP, +\textbf{\textcolor{brightgreen}{16.2}}\%). These results indicate that the proposed framework is more effective in generating precise and spatially consistent anomaly maps under challenging industrial environments. Although Dinomaly achieves a higher pixel-level AUROC, its lower AP and $F_1$ scores suggest that the generated anomaly maps contain less precise localization boundaries and more false-positive activations. Since pixel-level anomaly detection is inherently highly imbalanced, AUROC mainly reflects the global ranking ability between anomalous and normal pixels across all thresholds. Consequently, diffuse anomaly responses can still achieve strong AUROC performance if anomalous pixels generally receive higher scores than normal pixels. However, AP and $F_1$ are more sensitive to localization precision and false-positive predictions under the optimal dataset-level threshold. The higher AP and $F_1$ scores achieved by SwinAD therefore indicate that the proposed framework can better discriminate anomalous regions from normal regions while producing cleaner and more accurate segmentation masks.

\begin{table*}[t!]
\centering
\caption{Quantitative comparison on RealiAD. The table summarizes image-level classification performance and pixel-level anomaly localization performance across state-of-the-art methods. SwinAD achieves the strongest image-level scores and improves pixel-level AP and $F_1$ while using a lower input resolution than Dinomaly.}
\setlength{\tabcolsep}{4pt} 
\renewcommand{\arraystretch}{0.95} 
\small
\begin{tabular}{p{0.15\textwidth}
      >{\centering}p{0.1\textwidth}
      >{\centering}p{0.1\textwidth}
      >{\centering}p{0.1\textwidth}
      >{\centering}p{0.1\textwidth}
      >{\centering}p{0.1\textwidth}
      >{\centering}p{0.1\textwidth}
      >{\centering\arraybackslash}p{0.1\textwidth}
      }
\toprule
\multirow{2}{*}{Method} & \multirow{2}{*}{Image Size}
& \multicolumn{3}{c}{Image-level} 
& \multicolumn{3}{c}{Pixel-level} \\
\cmidrule(lr){3-5} \cmidrule(lr){6-8}
& & AUROC & AP & $F_1$ 
& AUROC & AP & $F_1$ \\
\midrule
UniAD~\cite{uniad}  & $256 \times 256$      & 83.0 & 80.9 & 74.3 & 97.3 & 21.1 & 29.2 \\
ReContrast \cite{guo2023recontrast} & $256 \times 256$ & 86.4 & 84.2 & 77.4 & 97.8 & 31.6 & 38.2 \\
DiAD~\cite{diad}  & $256 \times 256$        & 75.6 & 66.4 & 69.9 & 88.0 & 2.9 & 7.1 \\
ViTAD~\cite{vitad}    & $256 \times 256$    & 82.7 & 80.2 & 73.7 & 97.2 & 24.3 & 32.3 \\
MambaAD \cite{he2024mambaad} & $256 \times 256$   & 86.3 & 84.6 & 77.0 & 98.5 & 33.0 & 38.7 \\

Dinomaly \cite{dinomaly} & $392 \times 392$  & 89.3 & 86.8 & 80.2 
                          & \textbf{98.8} & 42.8 & 47.1 \\
\midrule
\textbf{SwinAD (ours)} & $256 \times 256$ & 
\textbf{89.5} \footnotesize (+\textbf{\textcolor{brightgreen}{0.2}}) & 
\textbf{87.5} \footnotesize (+\textbf{\textcolor{brightgreen}{0.7}}) & 
\textbf{80.4} \footnotesize (+\textbf{\textcolor{brightgreen}{0.2}}) & 
96.9 \footnotesize (-\textbf{\textcolor{red}{1.9}}) & 
\textbf{49.4} \footnotesize (+\textbf{\textcolor{brightgreen}{6.6}}) & 
\textbf{52.0} \footnotesize (+\textbf{\textcolor{brightgreen}{4.9}})  \\
\bottomrule

\end{tabular}
\vspace{-2mm}
\label{tab:realiad_results}
\end{table*}

This behavior is further validated by the qualitative results shown in Fig.~\ref{fig:visa_qualitative}. Dinomaly tends to produce broad anomaly activations around object boundaries and surrounding background regions, resulting in spatially diffuse responses that extend beyond the true defect areas. Such over-activated anomaly maps increase false-positive predictions and reduce boundary accuracy. In contrast, SwinAD generates more compact and structurally coherent anomaly responses that align more closely with the ground-truth masks. Additionally, the proposed framework effectively suppresses irrelevant background activations while preserving fine-grained defect structures, leading to improved localization precision and stronger pixel-level evaluation performance. Overall, SwinAD achieves more robust localization performance under complex real-world industrial scenarios, particularly when anomalous regions occupy only a very small portion of the image.

\noindent{\textbf{Evaluations on Real-IAD}}: The quantitative comparison on the Real-IAD dataset is presented in Table \ref{tab:realiad_results}. Real-IAD is a more realistic and challenging industrial anomaly detection benchmark containing substantial intra-class variations, cluttered backgrounds, illumination changes, and subtle defect patterns. Compared with MVTec AD and VisA, anomalies in Real-IAD are often more diverse, and visually similar to surrounding normal structures, making accurate localization significantly more difficult. 

Overall, the proposed SwinAD framework achieves the strongest image-level performance among the compared methods. Specifically, SwinAD obtains 89.5\% image-level AUROC, 87.5\% AP, and 80.5\% $F_1$-score, slightly outperforming Dinomaly despite operating at a lower input resolution of 256×256 compared with 392×392. These results demonstrate that the proposed hierarchical reconstruction framework maintains strong anomaly discrimination capability even under complex real-world industrial conditions while reducing the computational cost associated with higher-resolution inference.

\begin{figure}[pos=tbp]
    \centering
    \includegraphics[width=1.0\linewidth]{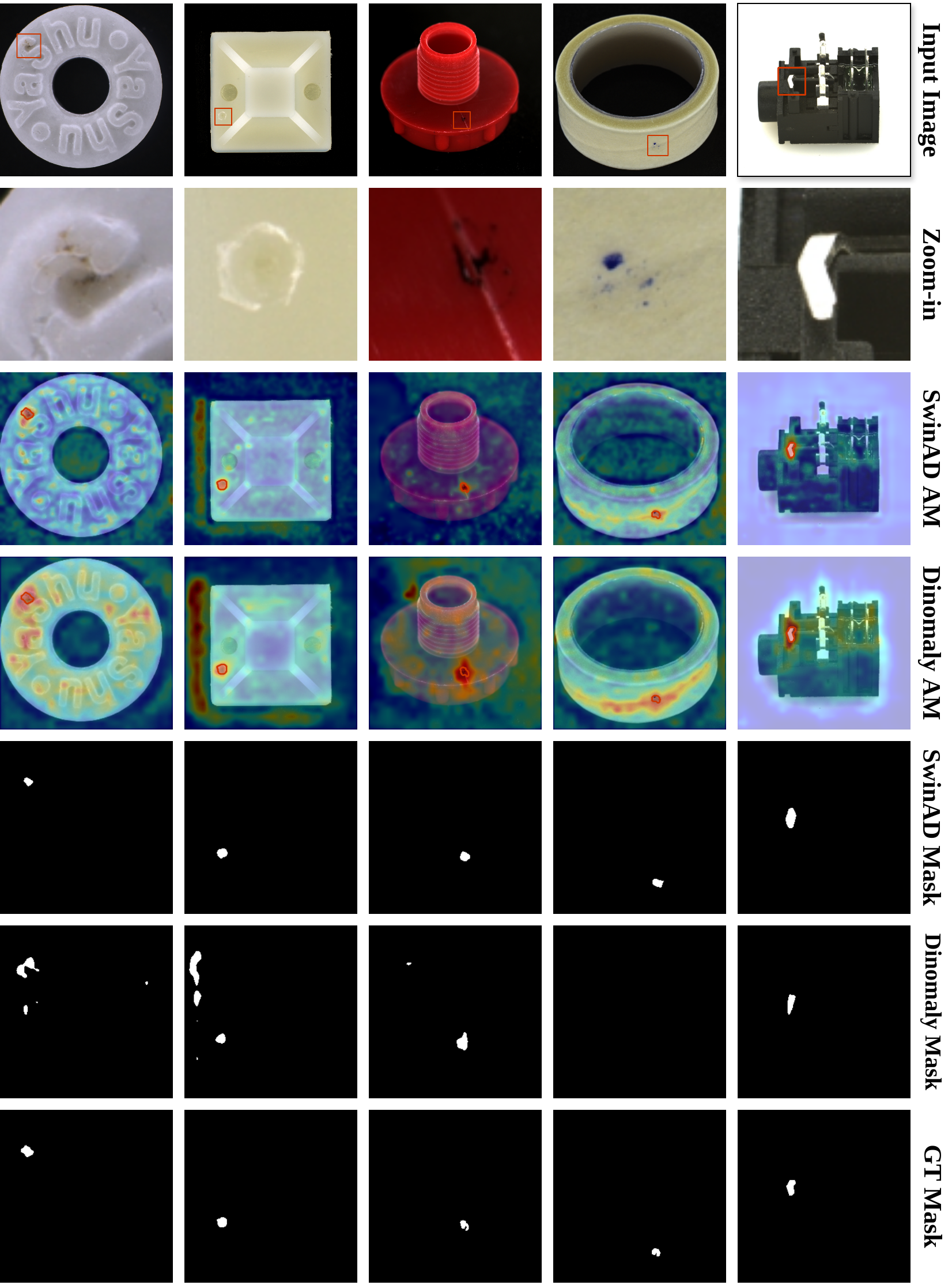}
    \caption{Qualitative comparison of anomaly localization on Real-IAD samples. From top to bottom, the figure presents input images, enlarged defect regions, anomaly maps from SwinAD and Dinomaly, the corresponding predicted masks, and ground-truth masks. Compared with Dinomaly, SwinAD better suppresses background responses and highlights defect regions with sharper spatial boundaries.}
    \label{fig:realiad_qualitative}
\end{figure}

At the pixel level, SwinAD achieves the best AP and $F_1$-scores, reaching 49.4\% AP and 52.0\% $F_1$, outperforming Dinomaly by \textbf{\textcolor{brightgreen}{6.6}}\% and \textbf{\textcolor{brightgreen}{4.9}}\%, respectively. Significant improvements are also observed compared with other recent methods such as ReContrast, ViTAD, and MambaAD. Although Dinomaly achieves a higher pixel-level AUROC of 98.8\%, its significantly lower AP and $F_1$-scores suggest that the generated anomaly maps contain broader and less precise activations. Similar to the observations on MVTec AD and VisA, anomalous regions in Real-IAD usually occupy only a small portion of the image. Under this highly imbalanced setting, diffuse anomaly responses can still achieve strong P-AUROC performance because anomalous pixels generally receive higher anomaly scores than normal pixels across most thresholds.

The qualitative results shown in Fig.~\ref{fig:realiad_qualitative} further support this observation. Dinomaly tends to generate spatially diffuse anomaly activations around defect boundaries and surrounding normal regions, resulting in over-expanded segmentation masks and increased false-positive responses. In contrast, SwinAD produces more compact and structurally coherent anomaly maps that better align with the ground-truth masks. The proposed framework effectively suppresses irrelevant background activations while preserving subtle defect structures and sharp spatial boundaries. Consequently, SwinAD achieves more robust localization performance in complex real-world industrial scenarios, particularly when anomalous regions occupy only a small portion of the image.

\begin{table}[pos=tbp]
    \centering
\caption{Computational efficiency comparison between SwinAD and Dinomaly.}
\setlength{\tabcolsep}{1pt} 
\renewcommand{\arraystretch}{1.25} 
\scriptsize
\begin{tabular}{p{0.07\textwidth}
      >{\centering}p{0.085\textwidth}
      >{\centering}p{0.085\textwidth}
      >{\centering}p{0.08\textwidth}
      >{\centering\arraybackslash}p{0.14\textwidth}
      }
\hline
\textbf{Method} & \textbf{Resolution} & \textbf{Params (M)} & \textbf{FLOPs (G)} & \textbf{Pixel AP} \\ \hline
Dinomaly & 224$\times$224 & 185.83 & 88.96 & 63.0/45.6/- \\
Dinomaly & 392$\times$392 & 185.83 & 268.92 & 69.3/53.2/42.8 \\
SwinAD	& 256$\times$256 & \textbf{159.01} & \textbf{54.32} & \textbf{74.4}/\textbf{58.0}/\textbf{49.4} \\ 

\bottomrule

\end{tabular}
\vspace{-2mm}
\label{tab:computational}
\end{table}
\textbf{Computational Efficiency Analysis}: To evaluate deployment practicality, we compare the computational complexity of the proposed framework with Dinomaly, the strongest recent transformer-based MUAD baseline. As shown in Table \ref{tab:computational}, SwinAD requires 159.01M parameters and 54.32 GFLOPs at an input resolution of 256$\times$256, whereas Dinomaly requires 185.83M parameters and 88.96 GFLOPs at 224$\times$224, increasing to 268.92 GFLOPs at 392$\times$392. Despite the significantly lower computational cost, SwinAD achieves higher pixel-level AP on all three benchmarks, improving from 69.3/53.2/42.8 obtained by Dinomaly to 74.4/58.0/49.4 on MVTec AD, VisA, and Real-IAD, respectively. In addition, SwinAD achieves an inference speed of approximately 20 FPS on single NVIDIA RTX 5000 GPU. These results indicate that the proposed hierarchical reconstruction framework provides a more favorable balance between localization accuracy and computational efficiency, making it more suitable for scalable industrial anomaly detection systems where both performance and inference cost are important considerations.

\textbf{Practical Deployment Considerations} 
Beyond quantitative performance, the proposed SwinAD framework is designed with practical industrial deployment. As discussed above, SwinAD achieves an inference speed of approximately 20 FPS on a single NVIDIA RTX 5000 GPU, making it suitable for real-time or near real-time inspection scenarios in manufacturing environments.

The anomaly map produced by SwinAD can be easily integrated into existing inspection systems. At the image level, the anomaly score can be used to generate alerts, reject defective products, or send suspicious samples for further inspection. At the pixel level, the anomaly map highlights the defect location, helping operators quickly identify abnormal regions and review potential defects. The anomaly map can also support downstream quality-control processes, such as defect analysis and inspection reporting. Because SwinAD provides both an image-level score and a pixel-level anomaly map, it can support fully automated inspection as well as human-in-the-loop quality assurance workflows.

\begin{table*}[t!]
\centering
\caption{Ablation study of stage-wise features and dual-decoder reconstruction on MVTec AD, VisA, and Real-IAD. Stage $i$ denote anomaly responses computed from $i$-th Swin encoder stage, Decoder 1 and Decoder 2 are the features from two reconstruction branches, and SwinAD indicates the full multi-stage aggregation.}
\setlength{\tabcolsep}{4pt} 
\renewcommand{\arraystretch}{0.95} 
\small
\begin{tabular}{p{0.125\textwidth}
      p{0.125\textwidth}
      >{\centering}p{0.1\textwidth}
      >{\centering}p{0.1\textwidth}
      >{\centering}p{0.1\textwidth}
      >{\centering}p{0.1\textwidth}
      >{\centering}p{0.1\textwidth}
      >{\centering\arraybackslash}p{0.1\textwidth}
      }
\toprule
\multirow{2}{*}{Dataset} & \multirow{2}{*}{Feature}
& \multicolumn{3}{c}{Image-level} 
& \multicolumn{3}{c}{Pixel-level} \\
\cmidrule(lr){3-5} \cmidrule(lr){6-8}
& & AUROC & AP & $F_1$ 
& AUROC & AP & $F_1$ \\
\midrule
\multirow{7}{*}{MVTecAD} 
& Stage 1 & 91.9 & 96.4 & 94.0 & 85.1 & 44.9 & 46.2 \\
& Stage 2 & 98.5 & 99.5 & 97.5 & 96.7 & 70.4 & 67.3 \\
& Stage 3 & 98.6 & 99.5 & 97.7 & 97.3 & 63.2 & 62.3 \\
& Stage 4 & 91.6 & 95.6 & 93.8 & 89.4 & 39.0 & 43.5 \\
\cmidrule(lr){2-8}
& Single Decoder & 99.3 & 99.8 & 98.6 & 97.9 & 73.0 & 69.5 \\
\cmidrule(lr){2-8}
& Decoder 1 & 99.3 & 99.8 & 98.5 & 98.0 & 73.1 & 69.7 \\
& Decoder 2 & 99.3 & 99.7 & 98.6 & 97.4 & 72.8 & 69.4 \\
\cmidrule(lr){2-8}
& SwinAD & \textbf{99.4} & \textbf{99.8} & \textbf{98.8} & 
\textbf{98.1} & \textbf{74.4} & \textbf{70.9}  \\
\midrule

\multirow{7}{*}{ViSA} 
& Stage 1 & 94.8 & 95.5 & 91.2 & 88.5 & 37.0 & 42.0 \\
& Stage 2 & 95.7 & 96.8 & 92.5 & 95.0 & 48.1 & 50.5 \\
& Stage 3 & 91.9 & 93.8 & 87.7 & 97.3 & 43.9 & 48.2 \\
& Stage 4 & 82.0 & 84.4 & 80.9 & 88.4 & 18.9 & 26.9 \\

\cmidrule(lr){2-8}
& Single Decoder & 97.2 & 97.5 & 94.2 & 95.8 & 55.6 & 55.8 \\
\cmidrule(lr){2-8}
& Decoder 1 & 97.3 & 97.8 & 94.4 & 96.4 & 53.5 & 54.2 \\
& Decoder 2 & 97.4 & 97.7 & 94.6 & 97.0 & 56.5 & 56.4 \\
\cmidrule(lr){2-8}
& SwinAD & \textbf{98.0} & \textbf{98.2} & \textbf{95.3} & 
\textbf{97.0} & \textbf{58.0} & \textbf{57.3}  \\
\midrule 

\multirow{7}{*}{RealiAD} 
& Stage 1 & 83.0 & 80.4 & 73.7 & 91.2 & 31.9 & 37.4 \\
& Stage 2 & 83.6 & 80.1 & 74.8 & 93.3 & 34.0 & 40.0 \\
& Stage 3 & 83.7 & 80.3 & 74.5 & 96.5 & 26.8 & 33.8 \\
& Stage 4 & 76.3 & 73.6 & 68.9 & 88.4 & 11.9 & 20.3 \\
\cmidrule(lr){2-8}
& Single Decoder & 88.2 & 85.5 & 78.7 & 97.0 & 46.6 & 50.8 \\
\cmidrule(lr){2-8}

& Decoder 1 & 88.2 & 85.5 & 78.8 & 96.8 & 44.0 & 50.6 \\
& Decoder 2 & 88.0 & 85.6 & 78.6 & 97.1 & 48.3 & 51.2 \\
\cmidrule(lr){2-8}
& SwinAD & \textbf{89.5} & \textbf{87.5} & \textbf{80.4} & 
\textbf{96.9} & \textbf{49.4} & \textbf{52.0}  \\

\bottomrule

\end{tabular}
\vspace{-2mm}
\label{tab:ablation_study}
\end{table*}

\subsection{Ablation Study}

Table \ref{tab:ablation_study} presents the ablation study of stage-wise anomaly representations and the proposed feature diversity-preserving reconstruction on the MVTec AD, VisA, and Real-IAD datasets. The results indicate that anomaly localization performance strongly depends on the semantic level of the extracted features, while the proposed dual-hypothesis formulation consistently improves both image-level and pixel-level anomaly detection performance across all datasets.

For stage-wise anomaly representations, shallow and intermediate encoder stages generally provide stronger localization capability because these features preserve more fine-grained local textures and structural details. In particular, intermediate stages maintain a better balance between local spatial information and higher-level semantic context. Unlike shallow stages that mainly capture low-level texture patterns, intermediate representations contain richer semantic understanding while still preserving sufficient spatial resolution for accurate localization. In contrast, deeper stages progressively focus on compact semantic representations and broader contextual structures, which may suppress subtle local anomaly cues due to semantic compression and progressive downsampling. Consequently, intermediate-stage features are more effective for detecting subtle structural inconsistencies and generating spatially precise anomaly maps across diverse industrial scenarios. This behavior can be clearly observed across all three datasets. On MVTec AD, Stage 2 achieves the best pixel-level AP of 70.4\% and an F1-score of 67.3\%, while Stage 3 achieves the highest pixel-level AUROC of 97.3\%. Similar trends are also observed on VisA and Real-IAD, where Stage 2 and Stage 3 consistently outperform both shallow and deep stages in most evaluation metrics. These observations suggest that intermediate representations provide the best trade-off between anomaly sensitivity and semantic robustness for industrial anomaly localization.

Although Stage 1 preserves the finest-grained local textures and structural details, its representations are also more sensitive to low-level noise, illumination variations, and background patterns. As a result, the extracted anomaly responses may contain redundant activations and weaker semantic discrimination, leading to lower overall localization performance compared with intermediate stages. In contrast, deeper stages mainly encode highly compressed semantic information and larger contextual structures. Although such representations are beneficial for image-level anomaly discrimination, they gradually lose fine spatial details that are important for precise localization. As a result, Stage 4 consistently produces weaker pixel-level performance, particularly on VisA and Real-IAD where anomalies are often small, sparse, and visually similar to surrounding normal regions. This degradation is especially noticeable in AP and F1-metrics, indicating that deeper semantic representations alone are insufficient for accurate anomaly boundary localization. Combining anomaly representations from all encoder stages further achieves significantly better performance by integrating fine-grained local textures with high-level semantic contextual information. The multi-stage aggregation enables the framework to simultaneously capture subtle local defects and broader structural inconsistencies, resulting in more robust anomaly discrimination and more spatially coherent localization across diverse industrial scenarios.

The ablation results also validate the effectiveness of the proposed feature diversity-preserving reconstruction framework. Individual decoder branches already achieve strong performance across all datasets, demonstrating that each reconstruction hypothesis can independently learn meaningful manifold-consistent normal representations. However, combining both reconstruction branches consistently provides additional improvements across both image-level and pixel-level metrics. For example, on MVTec AD, the proposed SwinAD framework improves pixel-level AP from 73.0\% obtained by the single decoder to 74.4\% after dual-hypothesis aggregation. Similar improvements can also be observed on VisA and Real-IAD, particularly for pixel-level AP and F1, which are more sensitive to localization precision.

\begin{figure*}[pos=t]
    \centering
    \includegraphics[width=1.0\textwidth]{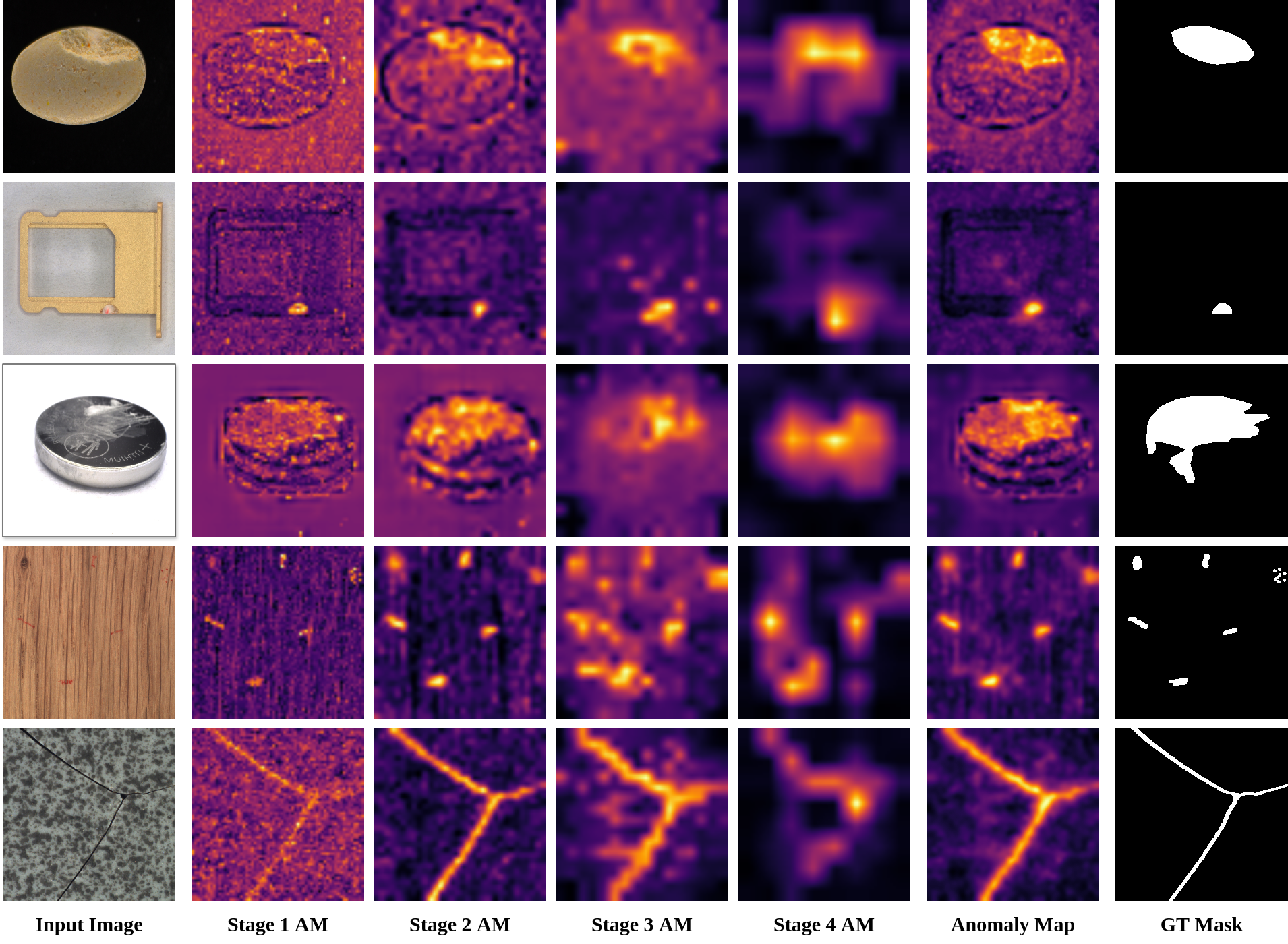}
    \caption{Visualization of the stage-wise ablation study. From left to right: Input image, anomaly map computed from individual stages, final anomaly map, and ground-truth anomaly mask. The figure compares anomaly responses obtained from individual Swin encoder stages and corresponding decoder features, and the full SwinAD model. While shallow stages emphasize local texture variations, deeper stages capture more semantic defect cues, and the complete multi-stage aggregation produces more accurate and spatially coherent anomaly localization.}
    \label{fig:stage_ablation_study}
    
\end{figure*}

These improvements indicate that the two reconstruction branches capture complementary manifold-consistent normal structures rather than learning redundant representations. By preserving multiple reconstruction hypotheses, the proposed framework improves manifold coverage and maintains greater feature diversity for normal patterns under varying appearances and structural variations. Consequently, normal regions remain more compatible with the jointly reconstructed feature manifold, while anomalous local structures become increasingly inconsistent with the learned normal representations. This behavior substantially improves anomaly discrimination capability and produces more spatially coherent anomaly maps.

Further analysis reveals that the benefits of the proposed feature diversity-preserving framework become more significant on more challenging datasets such as VisA and Real-IAD. These datasets exhibit larger intra-class variations, more complex backgrounds, illumination changes, and smaller anomalous regions, increasing the difficulty of distinguishing subtle defects from diverse normal structures. Under such conditions, deterministic single-branch reconstruction methods are more prone to over-generalization and may reconstruct anomalous regions as normal patterns. In contrast, the proposed feature diversity-preserving reconstruction framework maintains stronger sensitivity to subtle local inconsistencies while suppressing unnecessary activations on surrounding normal regions. As a result, SwinAD achieves more robust and precise anomaly localization performance under complex real-world industrial scenarios.

Figure \ref{fig:stage_ablation_study} presents qualitative anomaly localization results from different feature stages of the proposed framework. From left to right, each row shows the input image, anomaly maps generated from individual Swin encoder stages, the final aggregated anomaly map, and the corresponding ground-truth mask. The visualization illustrates how anomaly responses evolve across different semantic levels and how multi-stage aggregation improves localization quality.

The anomaly maps generated from shallow stages mainly emphasize fine-grained textures, edges, and local structural irregularities. In particular, Stage 1 produces high-resolution responses with strong sensitivity to subtle appearance variations and detailed structural boundaries. This behavior is especially beneficial for detecting tiny scratches, contamination, and thin crack patterns. However, because shallow features mainly encode low-level visual information, they are also more sensitive to background textures, illumination variations, and local noise. As a result, the generated anomaly maps often contain scattered activations and redundant responses on normal regions.

As the feature hierarchy becomes deeper, the anomaly responses gradually become more semantically concentrated and spatially stable. Intermediate stages provide a better balance between local structural details and higher-level semantic context. Compared with shallow stages, these representations suppress irrelevant background noise while still preserving sufficient spatial information for accurate localization. Consequently, Stage 2 and Stage 3 produce more coherent anomaly activations around defective regions and achieve stronger localization consistency across different anomaly types.

The deepest feature stage mainly focuses on high-level semantic inconsistency and global structural abnormalities. Although these representations provide stronger semantic confidence, progressive downsampling and semantic compression reduce fine spatial details and localization precision. As observed in the visualization, Stage 4 often generates coarse and spatially over-smoothed anomaly responses, making it less effective for accurately localizing small or thin defect regions. This limitation becomes particularly noticeable for subtle anomalies occupying only a small portion of the image.

Importantly, the anomaly maps generated from different stages exhibit complementary characteristics rather than redundant information. Shallow features preserve detailed local boundaries and texture sensitivity, while deeper features provide semantic robustness and contextual consistency. By aggregating anomaly evidence across multiple representation levels, the proposed framework effectively combines fine-grained structural information with high-level semantic understanding. This multi-scale integration substantially improves localization quality by reducing background noise, suppressing redundant activations, and enhancing spatial coherence around true defective regions.

As shown in the final aggregated anomaly maps, the proposed multi-stage reconstruction framework consistently localizes diverse anomaly types, including contamination, scratches, cracks, and irregular structural defects. Compared with individual stage-wise responses, the final anomaly maps exhibit sharper boundaries, more compact activations, and stronger alignment with the ground-truth masks. These results further demonstrate the effectiveness of the proposed manifold-consistent multi-scale anomaly estimation framework for robust and precise industrial anomaly localization.
\section{Conclusions}
\label{sec:conclusions}
In this paper, we presented \textbf{SwinAD}, a reconstruction-based framework for multi-class unsupervised anomaly detection. The proposed method leverages a frozen pretrained Swin Transformer V2 encoder to extract hierarchical multi-scale representations that preserve both local texture details and high-level semantic structures. Following the previous works, we adopt stage-wise bottleneck modules with dropout regularization to avoid trivial feature copying. Building upon this foundation, we introduce a feature diversity-preserving reconstruction decoder to generate complementary reconstruction hypotheses. By aggregating reconstruction discrepancies across multiple scales, the proposed SwinAD produces spatially coherent anomaly maps for accurate defect localization.

Extensive experiments on MVTec AD, VisA, and Real-IAD dataset demonstrate that SwinAD achieves competitive image-level anomaly detection performance and strong pixel-level localization accuracy, especially in terms of AP and $F_1$-score. The results show that hierarchical Swin features and feature diversity-preserving decoder reconstruction are effective for handling diverse normal patterns in the multi-class setting while maintaining efficient inference at a lower input resolution. In future work, we plan to explore adaptive scale weighting and more generalize manifold geomery reconstruction decoder designs to further improve localization robustness and efficiency in real-world industrial inspection scenarios.

{
    \small
    \bibliographystyle{ieeenat_fullname}
    \bibliography{main}
}


\end{document}